\documentclass{article} 
\usepackage[preprint]{colm2026_conference}

\usepackage{microtype}
\usepackage{hyperref}
\usepackage{url}
\usepackage{booktabs}
\usepackage{multirow}
\usepackage{multicol}
\usepackage{enumitem}
\usepackage{makecell}
\usepackage{wrapfig}
\usepackage{graphicx}
\usepackage{marvosym}
\usepackage{amsmath}
\usepackage{amssymb}
\usepackage{xcolor}
\usepackage{algorithm}
\usepackage{algpseudocode}
\usepackage{tcolorbox}
\tcbuselibrary{breakable}
\usepackage{CJKutf8}

\usepackage{tikz}
\usetikzlibrary{positioning}

\definecolor{darkblue}{rgb}{0, 0, 0.5}
\hypersetup{colorlinks=true, citecolor=darkblue, linkcolor=darkblue, urlcolor=darkblue}

\title{DuMate-DeepResearch: An Auditable Multi-Agent System with Recursive Search and Rubric-Grounded Reasoning}

\author{\bf DuMate Team, Baidu AI Cloud
}

\begin{document}

\ifcolmsubmission
\linenumbers
\fi

\maketitle

\begin{abstract}
Deep Research (DR) has emerged as a new agentic paradigm to tackle complex, open-ended research tasks, demanding systems that can iteratively frame problems, acquire evidence, verify sources, and synthesize long-form reports. 
In practice, however, current DR systems are constrained by four interrelated limitations: long-horizon planning over an underspecified scope, the bottleneck of decomposing and scheduling such tasks within a single agent, hallucination risk in long-form synthesis, and limited process auditability. This technical report presents \textbf{DuMate-DeepResearch}, a multi-agent DR framework built on the Qianfan Agent Foundry. The framework decouples the Agent Core---which handles task understanding, planning, and scheduling---from an extensible Tool Ecosystem for retrieval, evidence acquisition, and report rendering, making every intermediate decision and tool invocation explicitly traceable. Building on this infrastructure, DuMate-DeepResearch further introduces three mechanisms: (i) a \emph{graph-based dynamic planning} strategy expands the research roadmap coarse-to-fine and continuously revises it through reflection, re-planning, backtracking, and parallel branching; (ii) a \emph{recursive two-level execution} design delegates each complex search sub-task to an inner Search Agent that runs its own planning loop, isolating noisy retrieval and stabilizing long-horizon execution; (iii) a \emph{rubric-based test-time optimization} mechanism dynamically generates task-specific quality criteria and uses them as live reasoning scaffolds for evidence-grounded synthesis and adaptive stopping. 
Across two deep research benchmarks, DuMate-DeepResearch establishes new state-of-the-art results: the best overall score (58.03\%) on DeepResearch Bench, and the best overall score (61.95\%) on DeepResearch Bench II while ranking first in information recall and analysis. These results demonstrate the value of pairing auditable multi-agent infrastructure with adaptive planning and rubric-guided reasoning for high-quality deep research.
\end{abstract}

\section{Introduction}
\label{sec:intro}

The rapid advancement of artificial intelligence has catalyzed a paradigm shift from passive, single-turn question-answering systems to autonomous, agentic systems~\citep{yao_etal_2023_react,wang_etal_2024_survey}, enabling users to initiate complex research workflows from a research question. In this context, \textbf{Deep Research (DR)}~\citep{zheng_etal_2025_deepresearcher,shi_etal_2025_deep_research_systematic_survey,zhang_etal_2025_deep_research_survey_autonomous,du_etal_2025_deep_research_bench_comprehensive_benchmark,wang_etal_2025_live_research_bench_live_bench_mark_user_centric} has emerged as a crucial and highly challenging frontier to bridge the gap between human inquiry and systematic knowledge discovery.
While traditional retrieval-augmented workflows are confined to 
single-shot or rule-based retrieval over static corpora~\citep{lewis_etal_2020_retrieval,gao_etal_2023_retrieval}, DR aims to replicate the rigorous, systematic investigative methodologies of human researchers. To address complex, open-ended problems, DR requires sophisticated long-horizon reasoning, strategic decision-making, and large-scale information synthesis~\citep{shinn2023reflexion,yao2023tree}.

To operationalize such demanding workflows, recent efforts have explored a spectrum of architectural paradigms. Early systems adopted \textbf{monolithic architectures} (e.g., OpenAI's DeepResearch), which tightly integrate all modules around a central reasoning engine, ensuring unified control flow but limiting scalability and tool extensibility. Alternatively, \textbf{pipeline architectures} (e.g., n8n workflows) decompose the process into sequentially connected stages, facilitating component reuse but struggling with complex iteration and global feedback. In response, \textbf{agentic architectures} have become a natural direction for DR systems. By decomposing overarching research tasks and distributing them among autonomous agents with specialized roles, this collaborative paradigm improves scalability, parallel efficiency, and functional specialization for complex research scenarios.

\paragraph{Core Workflow of Deep Research}
Operating under this collaborative paradigm, the core workflow of modern agentic DR systems transcends a rigid linear pipeline, functioning instead as a closed-loop, tool-augmented process. Given a complex, open-ended research question, such a system transforms the high-level request into a comprehensive report through a set of tightly coupled capabilities that typically include, but are not limited to, the following:

\begin{enumerate}
    \item \textbf{Problem Framing and Adaptive Planning:} The system parses an underspecified research question into structured objectives and formulates a dynamic research roadmap, continuously revising its strategy as evidence accrues through sub-goal refinement, query reformulation, and backtracking from informational dead-ends.
    \item \textbf{Evidence Acquisition and Verification:} Driven by this roadmap, the system invokes a heterogeneous toolkit (e.g., web search engines, scholarly databases, domain-specific APIs) to acquire information, while assessing source credibility and cross-validating claims across sources to safeguard factual integrity.
    \item \textbf{Synthesis and Report Generation:} The validated evidence is finally integrated into a cohesive, logically structured report that weaves multi-source findings into a coherent narrative with nuanced analysis and verifiable citations.
\end{enumerate}

\paragraph{The Key Challenges}
However, realizing this idealized workflow in practice remains far from solved. Current agentic DR systems still confront open challenges that limit their reliability for real-world deployment:
\begin{itemize}
    \item \textbf{Long-Horizon Planning and Dynamic Scope Definition:} A research question unfolds into a long horizon of dozens of interdependent sub-questions whose scope is underspecified at the outset and only crystallizes as evidence accrues. Reactive, step-by-step policies that commit to a single next action---as in ReAct-style agents---are inherently myopic: they optimize locally without a global representation of the trajectory, oscillate between unbounded exploration and premature convergence, and cannot coherently revise their strategy when a tool fails or newly retrieved evidence invalidates an earlier premise. Effective DR therefore demands a planning formalism that maintains a global, far-sighted model of the entire roadmap and continuously re-delineates scope and re-plans as the information state evolves.

    \item \textbf{Complex Task Decomposition and Scheduling:} Even given a sound plan, decomposing and scheduling it for execution is where long trajectories most often break down. A single flat agent can rarely reconcile high-level task decomposition with the finer sub-task decomposition, scheduling, and noise handling that each sub-task in turn demands, since every sub-question may itself entail many multi-step retrieval actions over a stochastic web rife with dead links, API failures, and irrelevant or contradictory returns. Folding global strategy and low-level retrieval into one policy entangles the two and lets a single local failure propagate and cascade into the global trajectory. Reliable DR thus requires an execution scheme that separates high-level decomposition and scheduling from local sub-task completion, confines noise and errors within sub-task boundaries, and robustly carries out each sub-task without destabilizing the overall process.

    \item \textbf{Hallucination Mitigation and Factual Grounding:} Sustaining strict factual fidelity during long-form synthesis over dynamic, multi-source evidence streams is notoriously difficult, and the agent must additionally possess a principled criterion for when accumulated evidence is sufficient to halt exploration. This calls for rigorous inference-time scaffolds that calibrate every salient assertion against verifiable evidence as it is generated, and that terminate retrieval precisely when---and only when---the evidence demonstrably suffices, rather than relying on post-hoc verification or fixed exploration budgets.

    \item \textbf{Process Explainability and Auditability:} For DR to be trusted in high-stakes domains, its autonomous reasoning must be rendered inspectable. Systems should externalize their decision traces, tool invocations, and action paths as explicit, auditable artifacts---as transparent as the methodology appendix of a rigorous study---so that users can scrutinize not only the final report but the very process by which it was produced.
\end{itemize}

To address these challenges, we present \textbf{DuMate-DeepResearch}, an end-to-end multi-agent research framework. Built on top of the \textit{Qianfan Agent Foundry}, our system decouples the central cognitive brain (Agent Core) from the versatile execution layer (Tool Ecosystem).
This decoupling not only enables independent evolution of cognition and tooling, but also exposes every planning decision and tool invocation as an inspectable artifact, directly targeting the transparency and auditability challenge.
Furthermore, we equip the framework with three cognitive mechanisms tailored to DR:
(i) a \textbf{graph-based dynamic planner} that casts the research roadmap as an evolving directed acyclic graph, expanded coarse-to-fine and continuously revised through reflection, re-planning, backtracking, and parallel branching. Unlike myopic step-by-step ReAct-style reasoning, this graph maintains a global, far-sighted view of the entire trajectory and re-thinks its strategy whenever a tool fails or new evidence overturns an earlier assumption—jointly delivering long-horizon foresight and dynamic scope control;
(ii) a \textbf{recursive two-level execution} design, in which the outer Research Agent delegates every complex search sub-task to an inner \emph{Search Agent} that is itself a complete Foundry Agent running its own planning--execution cycle. This nesting isolates noisy, multi-step retrieval from high-level research strategy, so that a single failed search cannot destabilize the global trajectory—the key to stable execution under stochastic web conditions;
and (iii) a \textbf{rubric-based test-time optimization} mechanism that synthesizes question-specific evaluation rubrics dynamically and uses them as inference-time reasoning scaffolds to ground generated claims in retrieved evidence, while also providing an adaptive termination criterion.

We conduct extensive experiments on two deep research benchmarks. On DeepResearch Bench, DuMate-DeepResearch attains the best overall score among strong commercial and open baselines, establishing new state-of-the-art performance. On DeepResearch Bench II, which evaluates reports through fine-grained expert-derived rubrics, DuMate-DeepResearch also achieves the best overall score and leads on the information recall and analysis dimensions. Together, these results provide consistent evidence that the proposed architecture improves both broad report quality and rubric-grounded evidence acquisition and synthesis.

In summary, the main contributions of this report are summarized as follows:
\begin{itemize}
    \item \textbf{A decoupled multi-agent infrastructure for auditable DR:} We introduce the Qianfan Agent Foundry, a highly scalable architecture that implements a transparent \emph{understanding--planning--execution} cyclic paradigm by separating the reasoning core from the tool ecosystem, yielding a DR pipeline whose entire trajectory is auditable.

    \item \textbf{A graph-based dynamic planning algorithm:} We represent the research roadmap as a dynamic directed acyclic graph expanded in a coarse-to-fine manner and equipped with reflection, re-planning, backtracking, and parallel branching. In contrast to myopic ReAct-style reasoning that commits to one next action at a time, this graph sustains a global, far-sighted view of the trajectory and self-revises as evidence accumulates, jointly delivering long-horizon foresight and adaptive scope control.

    \item \textbf{A recursive two-level execution framework:} We instantiate the Foundry paradigm \emph{recursively}: the outer planning agent decomposes the deep-research task into sub-tasks, and each complex search sub-task is in turn solved by an inner search agent that is itself a complete Foundry Agent with its own planning--execution cycle. This nesting isolates noisy, multi-step retrieval from high-level strategy, preventing a single failed search from destabilizing the global trajectory and substantially improving execution stability.

    \item \textbf{Rubrics as test-time reasoning scaffolds:} We adapt dynamically generated rubrics from evaluation signals into inference-time scaffolds that calibrate generation against retrieved evidence, supporting factual grounding and bounding exploration through an adaptive stopping criterion.

    \item \textbf{State-of-the-art empirical performance:} We conduct extensive experiments on DeepResearch Bench and DeepResearch Bench II. The results demonstrate that DuMate-DeepResearch outperforms existing commercial and open baselines on both benchmarks, establishing new state-of-the-art performance across overall report quality, information recall, and analysis.
\end{itemize}

\section{DuMate-DeepResearch Framework}

DuMate-DeepResearch is an end-to-end Deep Research Agent built upon the Qianfan Agent Foundry. It follows an agentic loop of task understanding, planning, and execution to carry out complex, long-horizon research tasks.

\paragraph{Problem Formulation.}
{
DuMate-DeepResearch organizes each research session as an auditable, evidence-grounded state-transition process. Given a user query $q$, the Router produces a structured task specification; the Planner maintains an evolving research plan; the Execution Module invokes tools or Search Agents and accumulates evidence; and a rubric-guidance signal steers planning, stopping, and writing. This design allows the system to revise its research path while preserving the global report structure and the evidence trail.

We formalize this loop as a state-transition system over long-horizon research trajectories. At iteration $t$, the agent maintains a research state
\begin{equation}
s_t=\langle z,\; p_t,\; e_t,\; \rho_t\rangle,
\end{equation}
where $z=(x,\mathcal{O})$ is the fixed task context that bundles the research topic $x$ and the report outline $\mathcal{O}$; $p_t$ is the current research plan; $e_t$ is the accumulated evidence base collected from completed actions; and $\rho_t$ is the current guidance signal. Later subsections instantiate $p_t$ as a graph-structured plan (Section~\ref{sec:graphplan}) and $\rho_t$ as a rubric-based control signal (Section~\ref{sec:rubric}). The increment $\Delta e_t$ contains newly collected evidence lists and evidence summaries returned by direct tool actions or Search Agents, including source-grounded records and consolidated findings for executed sub-tasks; the global evidence base is their accumulation over cycles. Starting from $s_0=\langle z,p_0,\varnothing,\rho_0\rangle$, each cycle plans a set of executable actions $a_t$, executes them to obtain newly collected evidence $\Delta e_t$, and folds the new information and updated guidance back into the state,
\begin{equation}
s_{t+1}=\mathcal{T}\bigl(s_t,\,a_t,\,\Delta e_t\bigr).
\end{equation}
The loop continues until a stopping predicate $\textsc{Stop}(s_t)$ holds---for example, when the plan is fully explored or the current guidance signal reports no outstanding evidence gap---after which the Writer synthesizes the long-form report $y$ from the accumulated evidence. The three subsequent parts instantiate this loop: Section~\ref{sec:core} details the Router, Planner, and Execution modules; Section~\ref{sec:graphplan} specifies the graph-structured transition; and Section~\ref{sec:rubric} defines the rubric mechanism that implements the guidance signal. Algorithm~\ref{alg:agentloop} states the overall control loop.
}
\begin{figure}[htbp]
    \centering
    \includegraphics[width=\linewidth]{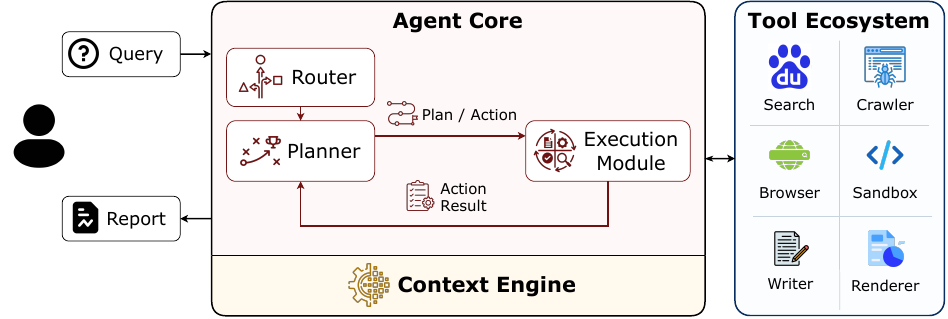}
    \caption{The illustration for the Qianfan Agent Foundry.}
    \label{fig:qianfan_agent_infra}
\end{figure}

\subsection{Qianfan Agent Foundry}

As a foundational infrastructure designed for general LLM-based agent construction, the Qianfan Agent Foundry consists of two decoupled components (illustrated in Figure~\ref{fig:qianfan_agent_infra}): the Agent Core and the Agent Extension (Tool Ecosystem). While the Agent Core functions as the central cognitive brain—orchestrating reasoning, planning, and task scheduling—the Agent Extension serves as the versatile execution layer. It provides a comprehensive suite of tools that empower the agent to interact with external environments, gather empirical evidence, and render final deliverables. This decoupled architecture ensures both robust cognitive control and highly extensible execution capabilities.

\begin{algorithm}[t]
\caption{DuMate-DeepResearch Agent Loop}
\label{alg:agentloop}
\begin{algorithmic}[1]
\Require user query $q$, max iterations $T_{\max}$
\State $x \gets \mathcal{U}(q)$ \Comment{Router: task understanding and analysis}
\State $\mathcal{O} \gets \textsc{Outline}(x, e_{t_c})$;\quad $z \gets (x,\mathcal{O})$ \Comment{Writer builds the outline from coarse-exploration evidence $e_{t_c}$; then fixed}
\State $p_0 \gets \textsc{InitPlan}(x,\mathcal{O})$;\quad $e_0 \gets \varnothing$;\quad $\rho_0 \gets \textsc{InitGuidance}(x,\mathcal{O})$
\State $s_0 \gets \langle z, p_0, e_0, \rho_0\rangle$;\quad $t \gets 0$
\While{$t \le T_{\max}$ \textbf{and} \textbf{not} $\textsc{Stop}(s_t)$}
   \State $a_t \gets \mathcal{P}(s_t)$ \Comment{Planner: graph-based dynamic planning}
   \State $\Delta e_t \gets \mathcal{X}(s_t,a_t)$ \Comment{Execution: evidence collection}
   \State $s_{t+1} \gets \mathcal{T}(s_t, a_t, \Delta e_t)$ \Comment{fold in evidence and updated guidance}
   \State $t \gets t+1$
\EndWhile
\State \Return $y \gets \mathcal{W}(x,\mathcal{O},e_t, \rho^{p})$ \Comment{Writer: guidance-conditioned synthesis}
\end{algorithmic}
\end{algorithm}

\subsubsection{DuMate-DeepResearch Core}
\label{sec:core}

The core of DuMate-DeepResearch comprises several specialized modules that collaborate seamlessly to effectively handle deep research tasks.

\paragraph{Router (Task Understanding and Analysis)}
The Router module is responsible for the initial comprehension and deconstruction of the research task. Given a user query, the Router extracts salient information and identifies the core research topic. This information is consolidated into a structured representation (e.g., a standardized JSON format), which is crucial for downstream planning and execution.
Furthermore, the Router serves as an intelligent interface for user interaction: if the initial query is ambiguous or incomplete, the Router proactively prompts the user for clarification. This design ensures that the research trajectory remains rigorously aligned with user expectations. In the global loop, the Router produces the topic specification $x$, and the Planner schedules the Writer to generate the outline $\mathcal{O}$; together they define the context $z=(x,\mathcal{O})$ for all downstream planning.

\paragraph{Planner (Task Thinking and Planning)}
The Planner module acts as the strategic engine, responsible for formulating the research methodology, reasoning through the investigative path, and planning future steps. Utilizing the structured task representation from the Router, the Planner analyzes the current knowledge state to identify critical epistemic gaps. It then strategically decomposes the overarching objective into tractable key research questions and actionable sub-problems. Based on this reasoning, the Planner selects the specific tools to be utilized and generates the corresponding parameters required for execution. Its graph-structured policy is developed in detail in Section~\ref{sec:graphplan}.

\paragraph{Execution Module (Planner-Following Task Scheduling and Execution)}
The Execution Module realizes the actions issued by the Planner, manages execution context, and collects the returned evidence; unlike the Router and Planner, it sets no research strategy of its own. Depending on the action type, it routes execution to one of four targets: a \emph{direct tool call}, whose interface it invokes and whose output it normalizes; a \emph{Search Agent}, dispatched for open-ended retrieval sub-tasks and itself a Foundry Agent with a local planning loop (Section~\ref{sec:hierarchy}) rather than a single black-box query; the \emph{Writer}, a generation agent invoked with two prompts---an outline prompt that turns the early coarse-exploration evidence into the fixed outline $\mathcal{O}$, and a report prompt that synthesizes the accumulated evidence into the final long-form report; and a \emph{lightweight reasoning} (\emph{llm}) action that deduplicates, merges, and cross-validates collected evidence without issuing new retrieval. Supporting serial and parallel fan-out across these targets, it acts as a scheduling and dispatch layer that carries out the Planner's decisions while leaving every high-level research choice to the Planner.

The collaboration among these modules makes the research trajectory explicitly inspectable. The Router maintains the structured task representation, the Planner records decision traces and sub-task decompositions, and the Execution Module logs tool invocations and retrieved evidence. As a result, users can inspect not only the final report but also the intermediate reasoning and action paths that produced it.

\subsubsection{DuMate-DeepResearch Extension: The Tool Ecosystem}
Complementing the cognitive core, DuMate-DeepResearch integrates a comprehensive Tool Ecosystem. Driven by the Execution Module, this ecosystem serves as the versatile execution layer for the "task scheduling and execution" phase, encompassing diverse tools for information retrieval, data analysis, and report generation.

These tools are seamlessly integrated into the agentic execution framework, allowing for efficient coordination and utilization throughout the research process. By leveraging this tool ecosystem, DuMate-DeepResearch can effectively handle the diverse and complex requirements of deep research tasks, further enhancing its capabilities and performance in delivering high-quality research outcomes. We introduce two key tools in DuMate-DeepResearch's tool ecosystem as follows.

\paragraph{Baidu Search Integration}
Baidu Search provides the primary retrieval substrate for evidence acquisition in DuMate-DeepResearch. Rather than treating search as a single black-box query, the Execution Module exposes retrieval as a set of structured actions, including query expansion, web search, direct URL crawling, page-content extraction, and evidence normalization. Returned snippets and pages are converted into evidence records that preserve source metadata, URLs, timestamps when available, and short summaries for downstream verification and citation-aware synthesis. This design separates retrieval infrastructure from research policy: the Planner and Search Agents decide what information is needed and how queries should evolve, while the Tool Ecosystem supplies traceable evidence for cross-source checking and final report grounding.

\paragraph{Report Rendering Tools}
To ensure the high quality and formatting diversity of the final deliverables, DuMate-DeepResearch employs a decoupled, two-stage report rendering mechanism. Initially, the system generates a unified "pivot report," utilizing robust reasoning capabilities to guarantee logical coherence and content comprehensiveness. Subsequently, specialized rendering tools translate this pivot report into multiple user-desired formats (e.g., Markdown, HTML, PPT), ensuring adaptability across various presentation contexts.

\begin{figure}[t]
    \centering
    \includegraphics[width=\linewidth]{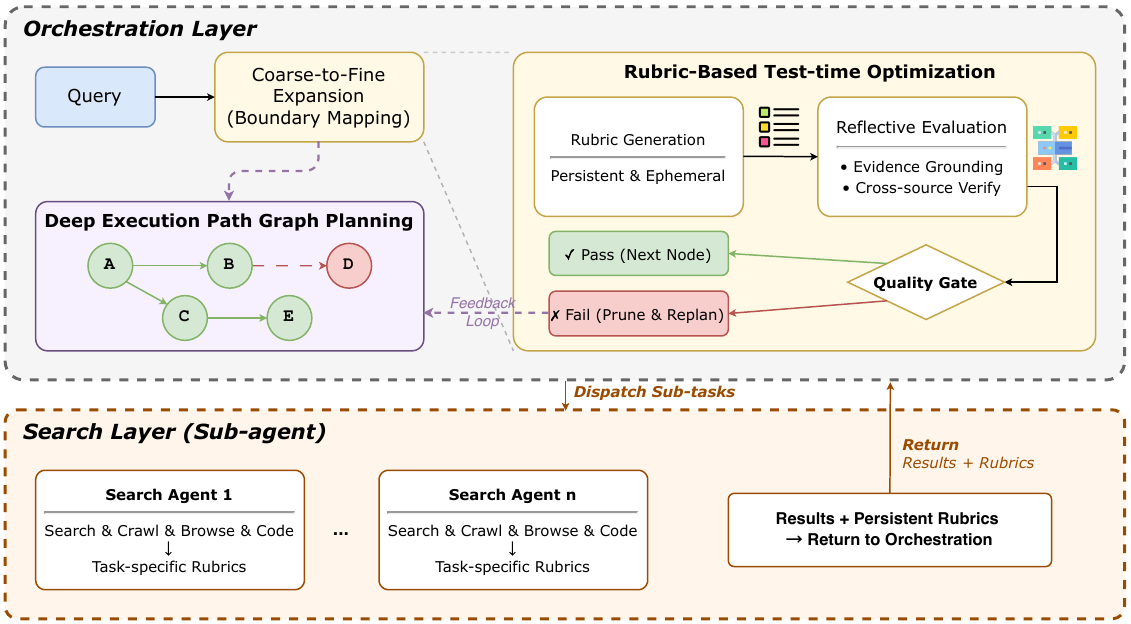}
    \caption{The illustration for dynamic planning and test-time optimization.}
    \label{fig:plan_and_rubric}
\end{figure}

\subsection{Dynamic Planning and Test-Time Optimization}
\label{sec:dpto}

On top of the Foundry infrastructure, DuMate-DeepResearch introduces three mechanisms that shape the long-horizon research process. First, \emph{graph-based dynamic planning} (Section~\ref{sec:graphplan}) rewrites the evolving plan as evidence accumulates, maintaining a global, self-revising roadmap instead of committing to a single next-action chain. Second, \emph{recursive two-level execution} (Section~\ref{sec:hierarchy}) lets the outer Research Agent delegate complex search sub-tasks to inner Search Agents that run their own local Foundry cycles, keeping noisy retrieval separate from high-level research strategy. Third, \emph{rubric-based test-time optimization} (Section~\ref{sec:rubric}) turns the guidance signal into active rubric instructions for planning, retrieval, stopping, and final synthesis. We develop the three mechanisms in turn, introducing notation only where it sharpens the mechanism being described (as shown in Figure~\ref{fig:plan_and_rubric}).

\subsubsection{Graph-Based Dynamic Planning}
\label{sec:graphplan}

\begin{wrapfigure}{r}{0.5\textwidth}
    \centering
    \includegraphics[width=0.48\textwidth]{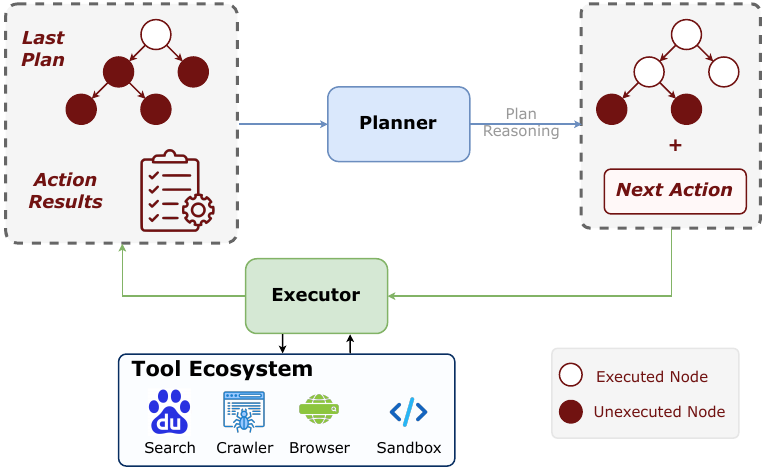}
    \caption{The illustration of deep execution path graph planning and reflection.}
    \label{fig:dumate_deep_research_plan}
\end{wrapfigure}
\paragraph{Coarse-to-Fine Expansion for Dynamic Scope}
DuMate-DeepResearch expands the research path in a coarse-to-fine manner. Complex tasks often begin with vague intent, making it difficult to balance broad exploration with premature convergence. The system therefore starts with a macro-level exploratory retrieval phase that maps the research space and establishes a preliminary cognitive framework. 
We use $t_c$ to denote the checkpoint at which this initial coarse-exploration phase completes; the corresponding evidence base $e_{t_c}$ is used by the Writer to construct the fixed outline $\mathcal{O}$ in Algorithm~\ref{alg:agentloop}.
Guided by the graph-based dynamic planner, the system then transitions to a granular phase, systematically diving into defined sub-topics to collect targeted evidence. This progressive decomposition and integration mechanism refines the research scope as evidence accumulates, calibrating the boundary between breadth and depth without losing focus.
We formalize this roadmap at planning iteration $t$ as a DAG-structured plan $p_t=(V_t,E_t)$, the planning component of the global state $s_t$ in Algorithm~\ref{alg:agentloop}. Each node $v\in V_t$ is a sub-task carrying a tuple $\langle d(v),\,\chi(v)\rangle$, where $d(v)\in\mathbb{Z}^{+}$ is its depth in the coarse-to-fine hierarchy (smaller values denote broader, exploratory sub-tasks), and $\chi(v)\in\{0,1\}$ is a binary execution status; a directed edge $(u,v)\in E_t$ records that $v$ depends on $u$. The coarse-to-fine principle then becomes a depth-ordered expansion in which the scheduler only ever dispatches the \emph{ready frontier},
\begin{equation}
\mathcal{F}_t=\bigl\{\,v\in V_t \;:\; \chi(v)=0 \;\wedge\; \forall (u,v)\in E_t,\; \chi(u)=1\,\bigr\},
\end{equation}
i.e.\ the unexecuted sub-tasks whose dependencies are all satisfied. Confining execution to $\mathcal{F}_t$ guarantees that broad, low-depth probes are resolved before their finer descendants are instantiated, so that boundary definition reduces to a monotone, dependency-respecting expansion rather than an unbounded search.

\paragraph{Far-Sighted Re-Planning over a Dynamic Graph}
The dynamic graph also gives the Planner a global structure for revising its strategy as evidence arrives (as shown in Figure~\ref{fig:dumate_deep_research_plan}). Myopic, step-by-step ReAct-style reasoning commits to one next action at a time and lacks a global view of the trajectory; in highly stochastic web environments it can stall on dead links, API errors, or contradictory evidence. Representing the roadmap as a dynamic graph instead gives the Planner a far-sighted view of the entire trajectory: at each milestone (node) the agent evaluates intermediate outcomes against expectations, and when anomalies surface it prunes dead ends, adjusts subsequent strategy, and re-plans alternative paths rather than greedily extending a single chain. This graph-level re-planning lets the system revise earlier assumptions whenever a tool fails or new evidence overturns them, yielding resilience over long horizons.
Formally, at each iteration the Planner emits a set of parallel actions $a_t$ over the ready frontier, where each action $\alpha\in a_t$ binds a frontier sub-task $v\in\mathcal{F}_t$ to a tool and its parameters. Once the Execution Module returns the newly collected evidence $\Delta e_t$ and folds it into the accumulated evidence base $e_{t+1}$, the roadmap is regenerated by a single re-planning operator
\begin{equation}
p_{t+1}=\Pi\bigl(p_t,\,e_{t+1},\,\rho_{t+1}\bigr),
\end{equation}
which updates only the plan component of the global state; the full transition additionally folds in the fresh evidence and updated guidance. Conditioned on the current plan, the accumulated evidence, and the latest guidance signal $\rho_{t+1}$, $\Pi$ may \emph{expand} the frontier with finer sub-tasks, \emph{prune} unproductive branches---backtracking away from dead links or contradictory evidence---or \emph{rewire} dependencies, while it always preserves executed nodes so that $\chi(v){=}1$ is monotone and no evidence is recomputed. To curb error propagation, every candidate action first passes a lightweight reflection gate before any tool is invoked; rejected actions are revised under the critic's feedback for a bounded number of rounds. The loop halts and yields to report synthesis once the frontier is exhausted ($\mathcal{F}_t=\varnothing$) or the Planner emits a terminal synthesis action, under a hard iteration bound $t\le T_{\max}$---exactly the stopping predicate of the global loop. Casting expansion, reflective re-planning, and adaptive stopping as the single operator $\Pi$ turns the long-horizon trajectory into one auditable update rule, summarized in Algorithm~\ref{alg:planning}.

\begin{algorithm}[t]
\caption{Graph-Based Dynamic Planning with Reflection}
\label{alg:planning}
\begin{algorithmic}[1]
\Require research topic $x$, report outline $\mathcal{O}$, max iterations $T_{\max}$
\State $p_0 \gets \textsc{InitPlan}(x,\mathcal{O})$;\quad $e_0 \gets \varnothing$;\quad $\rho_0 \gets \textsc{InitGuidance}(x,\mathcal{O})$;\quad $t \gets 0$
\While{$t \le T_{\max}$}
  \State $\mathcal{F}_t \gets \{\,v \in V_t : \chi(v){=}0 \wedge \text{deps}(v)\ \text{satisfied}\,\}$ \Comment{ready frontier}
  \State $a_t \gets \textsc{Planner}(p_t, \mathcal{O}, e_t, \rho_t)$ restricted to $\mathcal{F}_t$ \Comment{select parallel actions}
  \If{$a_t = \varnothing$ \textbf{or} $a_t$ is a synthesis action}
     \State \textbf{break} \Comment{adaptive stopping}
  \EndIf
  \While{reflection gate returns \textsc{revise} for $a_t$ \textbf{and} bounded rounds not reached}
     \State revise $a_t$ under critic feedback
  \EndWhile
  \State $\Delta e_t \gets \textsc{ExecuteParallel}(a_t)$ \Comment{via tools or bounded Search Agent dispatch}
  \State update $\chi(\cdot)$ for executed nodes
  \State $e_{t+1} \gets e_t \cup \Delta e_t$ \Comment{accumulate evidence for subsequent planning}
  \State $\rho_{t+1} \gets \textsc{UpdateGuidance}(\mathcal{O},e_{t+1})$
  \State $p_{t+1} \gets \Pi(p_t, e_{t+1}, \rho_{t+1})$ \Comment{expand / prune / rewire}
  \State $t \gets t + 1$
\EndWhile
\State \Return $y \gets \mathcal{W}\bigl(x,\mathcal{O}, e_t,\ \rho^{p}\bigr)$ \Comment{Writer: guidance-conditioned synthesis}
\end{algorithmic}
\end{algorithm}

A desensitized excerpt of the actual planner prompt that drives this procedure---retaining its DAG legality, depth-bounding, and re-planning constraints while omitting the output schema and other sensitive details---is provided in Appendix~\ref{app:planner-prompt}.

\subsubsection{Recursive Two-Level Execution}
\label{sec:hierarchy}

Even with a sound graph-based plan, execution remains difficult because each open-ended sub-task may itself require many noisy, multi-step retrieval actions. Folding high-level strategy and local search into one flat agent lets a single failed retrieval cascade into the global trajectory. DuMate-DeepResearch instead applies the Qianfan Agent Foundry \emph{recursively}, instantiating the same Router--Planner--Execution cycle at two nested levels with a clean division of labor.

At the \emph{outer} level, the Research Agent owns the global state $s_t$ and the plan $p_t$: it decides \emph{what to research} next and advances the research-planning loop of Algorithm~\ref{alg:agentloop}. Whenever a planned action is an open-ended retrieval sub-task, the outer Execution Module does not call a search tool directly; it dispatches an \emph{inner} Search Agent. Crucially, this agent follows the same Foundry abstraction---with its own Router, Planner, and Execution Module---but operates over a local search state for a single sub-task. It decides \emph{how to search}: formulating and reformulating queries, invoking the retrieval tools of the Tool Ecosystem, and consolidating the returned evidence until that sub-task is sufficiently covered, then returns evidence lists and summaries that are appended to the current cycle's $\Delta e_t$.

We capture this nesting with a compact level-indexed notation. Let $\mathcal{A}^{(\ell)}(q)$ denote a complete Foundry Agent that solves query $q$ at nesting level $\ell\in\{0,1\}$ and returns evidence lists and summaries; the outer Research Agent is $\mathcal{A}^{(0)}$. Applied to an open-ended retrieval action $a_v$ targeting sub-task $v$, the outer execution step instantiates an inner Agent on the sub-task query $q(v)$ one level down and folds the returned evidence into $\Delta e_t$.
The inner Agent $\mathcal{A}^{(1)}$ unfolds into the same Router--Planner--Execution cycle, subject to a single restriction that bounds the recursion: at the inner level, execution invokes the retrieval tools of the Tool Ecosystem directly rather than dispatching a further Agent. The nesting is therefore exactly two levels deep and terminates by construction, while the same execution abstraction appears at both levels, which is exactly what lets a complex search be carried out without conflating it with high-level planning.

The research process therefore unfolds as two nested loops---an outer research-planning loop wrapped around many parallel inner search loops---rather than one flat trajectory, and it is this recursion that stabilizes execution. It \emph{isolates failure}: a stalled or unproductive search is contained within a single Search Agent and cannot derail the global plan, while the outer Research Agent simply re-dispatches or re-plans around it. It \emph{separates concerns}: the outer Planner reasons over a compact graph of sub-tasks while each inner Agent reasons only within its own sub-task, so neither conflates strategy with search nor confronts the full combinatorial horizon. And because every level logs its own understanding--planning--execution trace, the recursive decomposition remains inspectable end to end.

\subsubsection{Rubric-Based Test-Time Optimization}
\label{sec:rubric}

\paragraph{From Evaluation to Reasoning Scaffold}
DuMate-DeepResearch further uses rubrics as test-time guidance for planning and synthesis. The concept of a rubric originates from long-form output evaluation. In standard RLVR (Reinforcement Learning with Verifiable Rewards), reward signals are typically binary, which is too coarse for open-ended report generation. Rubrics provide a more structured alternative by decomposing quality into fine-grained criteria such as evidence grounding, logical coherence, and multi-source cross-validation. Rather than using rubrics only as post-hoc evaluators, we inject them into the agents' reasoning process. This turns the rubric into a live scaffold that provides explicit criteria for source calibration and evidence-grounded synthesis.
We make this shift precise. A rubric is a set of criteria $\rho=\{c_1,\dots,c_k\}$ in which each criterion $c=\langle \text{name},\text{description},\text{guidance}\rangle$ has its \emph{guidance} field phrased as an actionable reasoning instruction rather than a numeric score. Whereas a conventional evaluator consumes a finished report and emits a scalar reward post hoc, we inject rubric context into generation itself before outputs are produced, compelling the agent to ground claims as it reasons rather than to be penalized afterward.

\paragraph{Dynamic Rubric Generation}
Because deep research is an evolving process, the rubric cannot remain entirely static. While the research goal is fixed, the information state changes as new evidence accumulates; criteria specified at initialization may become incomplete or misaligned with the current frontier. We therefore generate and update rubrics iteratively conditioned on the accumulated knowledge. The system uses two types of rubrics: \textit{Persistent Rubrics}, which define stable, topic-level quality dimensions applied uniformly across the session; and \textit{Ephemeral Rubrics}, which capture transient criteria derived from the latest retrieved information.
Let $\rho^{p}$ denote the persistent rubric and $\rho^{e}_t$ denote the ephemeral rubric available at cycle $t$. Concretely, the rubric-guidance signal $\rho_t$ introduced in Algorithms~\ref{alg:agentloop} and~\ref{alg:planning} is instantiated as an active rubric,
\begin{equation}
\rho_t=(\rho^{p},\rho^{e}_t).
\end{equation}
The initialization operator $\textsc{InitGuidance}$ first generates the persistent rubric from the research topic and the report outline,
\begin{equation}
\rho^{p}=\mathcal{G}_p(x,\mathcal{O}), \qquad \rho^{e}_0=\varnothing,
\end{equation}
where $\rho^{p}$ is then held fixed to anchor stable, topic-level quality dimensions. The update operator $\textsc{UpdateGuidance}$ refreshes the ephemeral rubric at the end of every cycle for use in the next,
\begin{equation}
\rho^{e}_{t+1}=\mathcal{G}_e(\mathcal{O},e_{t+1}), \qquad
\rho_{t+1}=(\rho^{p},\rho^{e}_{t+1}),
\end{equation}
conditioned on the accumulated evidence base $e_{t+1}$, so as to target the most decision-relevant gaps exposed by the current evidence state and track the moving information frontier in lockstep with the evolving plan. Under this instantiation, the Writer consumes the persistent component $\rho^{p}$ for final synthesis, while the Planner and Search Agents use the full active rubric during iterative research:
\begin{equation}
a_t \sim \pi_{\mathcal{P}}\bigl(\,\cdot \mid x,\,\mathcal{O},\,p_t,\,e_t,\,\rho^{p},\,\rho^{e}_t\,\bigr),
\qquad
y \sim \pi_{\mathcal{W}}\bigl(\,\cdot \mid x,\,\mathcal{O},\,e_t,\,\rho^{p}\,\bigr),
\end{equation}
where $a_t$ is the Planner action at cycle $t$, $y$ is the final long-form report, $\pi_{\mathcal{P}}$ and $\pi_{\mathcal{W}}$ denote the Planner and Writer policies, and $(x,\mathcal{O},p_t,e_t)$ is the current task context: the topic, fixed report outline, evolving plan, and accumulated evidence. The active rubric components thereby cease to be graders and become a \emph{live scaffold} for planning, while the persistent component provides the stable report-stage scaffold for prose generation.

\paragraph{Rubrics in Multi-Agent Collaboration}
Since DuMate-DeepResearch orchestrates the Agent Core and dispatched Search Agents in a hierarchical manner, with each level serving distinct objectives, the rubric strategy is designed accordingly. At the orchestration level, the active rubric $(\rho^{p},\rho^{e}_t)$ is refreshed after each planning-execution cycle and provided to the Planner for subsequent research decisions. At the search level, each Search Agent also receives active rubric guidance conditioned on its sub-task context and returned tool evidence. By contrast, the Writer consumes only the persistent report-stage rubric $\rho^{p}$ during final synthesis, so that dynamic evidence-gap guidance steers research control without becoming an additional moving constraint on report writing. Upon completing its search, the Search Agent returns evidence lists and summaries to the orchestration level, where they are incorporated into the accumulated evidence base.
This upward evidence flow closes the loop: the orchestrator feeds the updated evidence base into the next ephemeral rubric, so that the orchestration-level rubric stays aligned with what the search level actually uncovered. Crucially, the refreshed ephemeral rubric $\rho^{e}_{t+1}$ also serves as the adaptive termination signal: once it reports no outstanding gap, the stopping predicate $\textsc{Stop}$ of Algorithm~\ref{alg:agentloop} halts the loop, tying factual sufficiency directly to the stopping rule. Algorithm~\ref{alg:rubric} summarizes a single rubric-scaffolded reasoning step.

\begin{algorithm}[t]
\caption{Rubric-Scaffolded Test-Time Reasoning}
\label{alg:rubric}
\begin{algorithmic}[1]
\Require topic $x$, report outline $\mathcal{O}$, plan $p_t$, evidence base $e_t$, newly collected evidence $\Delta e_t$, persistent rubric $\rho^{p}$, active ephemeral rubric $\rho^{e}_t$ ($\rho^{e}_0=\varnothing$)
\State inject $\rho^{p},\rho^{e}_t$ into the Planner / Search Agent context and $\rho^{p}$ into the Writer context
\State generate Planner action $a_t$ conditioned on the active rubric
\State during synthesis, generate the Writer's report $y$ conditioned on the persistent rubric
\State $\rho^{e}_{t+1} \gets \mathcal{G}_e(\mathcal{O}, e_t \cup \Delta e_t)$ \Comment{ephemeral rubric: refreshed for the next cycle}
\If{$\rho^{e}_{t+1}$ reports no outstanding gap \textbf{or} reach max plan iteration} 
   \State signal \emph{stop} to the Planner \Comment{adaptive termination}
\EndIf
\State \Return $a_t$ during planning or $y$ during synthesis, together with $\rho^{p},\rho^{e}_{t+1}$
\end{algorithmic}
\end{algorithm}

Desensitized excerpts of the two-level rubric-generation prompts---both the orchestration-level prompt and the search-level prompt---are provided in Appendix~\ref{app:rubric-prompt}. They elicit the two rubric types, constrain the ephemeral criteria to the most decision-relevant evidence gaps, require the \emph{guidance} of each criterion to be an actionable instruction rather than a numeric score, and ask the generator to flag when no further retrieval is warranted, which yields the adaptive stopping signal.

\section{Experiments and Evaluation}

To assess the performance of the DuMate-DeepResearch system, we conducted comprehensive experiments on two deep research benchmarks:
\begin{itemize}
    \item \textbf{DeepResearch Bench}~\citep{du_etal_2025_deep_research_bench_comprehensive_benchmark}: A comprehensive benchmark specifically designed for deep research agents or systems. It includes a total of 100 tasks across 22 domains in both Chinese and English. The generated report for each task is evaluated using the Reference-based and Adaptive Criteria-driven Evaluation framework, which leverages LLM-as-a-judge for evaluation.
    \item \textbf{DeepResearch Bench II}~\citep{li_etal_2026_deep_research_bench_ii_diagnosing}: An extension of DeepResearch Bench, focusing on diagnosing deep research agents via rubrics derived from expert reports. It includes 132 tasks across 22 domains, with a total of 9,430 fine-grained binary rubrics for evaluation. The evaluation is conducted in an end-to-end manner, assessing the dimensions of Information Recall, Analysis, and Presentation.
\end{itemize}

\paragraph{Implementation Details}
Key hyperparameters are set as follows: the outer planning loop runs up to 15 iterations; each inner Search Agent performs up to 10 retrieval rounds, generating up to 3 sub-queries per round with 3 results returned per query; fan-out parallel execution is enabled so that independent sub-tasks on the ready frontier execute concurrently. Baidu Search serves as the primary retrieval backend. To account for variance in generation, all reported results for DuMate-DeepResearch are averaged over 3 independent runs.

\paragraph{Evaluation Protocol}
For both benchmarks, baseline scores are taken from the official benchmark sources and leaderboards, and DuMate-DeepResearch is evaluated under the corresponding official evaluation protocols. During report generation, the system is given only the benchmark queries and does not access benchmark reference reports, expert reports, or evaluation rubrics. This is particularly important for DeepResearch Bench II, whose evaluation rubrics are derived from expert reports; the rubrics generated by DuMate-DeepResearch are produced independently at test time and are not derived from the benchmark's hidden evaluation rubrics.

\subsection{Overall Performance}
\begin{table}[ht]
\centering
\resizebox{\textwidth}{!}{
\begin{tabular}{cccccc}
\toprule
Model/System & \textsc{Comprehensiveness} & \textsc{Insight} & \textsc{Instruction Following} & \textsc{Readability} & \textsc{Overall} \\
\midrule
DR-Tulu & 44.08 & 44.65 & 49.56 & 42.30  & 45.49 \\
UESTC-MBSE-RAAA & 43.77 & 48.34 & 47.21 & 43.78 & 46.13 \\
OpenAI DeepResearch* & 46.46 & 43.73 & 49.39 & 47.22 & 46.45 \\
Gemini 2.5 Pro DeepResearch* & 49.51 & 49.45 & 50.12 & 50.00    & 49.71 \\
\makecell{LangChain Open Deep Research\\(GPT-5 + Gensee Search)} & 50.06 & 50.76 & 51.31 & 49.72 & 50.60 \\
Salesforce AIR & 50.00    & 51.09 & 50.77 & 50.32 & 50.65 \\
ThinkDepth.ai & 52.02 & 53.88 & 52.04 & 50.12 & 52.43 \\
Tavily Research & 52.84 & 53.59 & 51.92 & 49.21 & 52.44 \\
LiAuto Mind DeepResearch 1.5 & 51.54 & 55.30  & 50.45 & 51.26 & 52.54 \\
RecallRadar Intelligence & 53.91 & 53.53 & 52.18 & 52.38 & 53.19 \\
Deep Dog 1 & 53.14 & 56.10  & 51.83 & 51.18 & 53.52 \\
Bodhi Deep Research & 54.23 & 56.09 & 52.86 & 51.81 & 54.22 \\
Onyx Deep Research & 54.67 & 56.43 & 53.08 & 52.02 & 54.54 \\
TrajectoryKit & 54.10  & 57.90  & 52.91 & 52.72 & 54.92 \\
CMCC-DeepInsight & 55.66 & 58.70  & 52.53 & 50.94 & 55.24 \\
MS-Agent DeepResearch & 56.76 & 56.79 & 53.10  & 52.28 & 55.31 \\
Cellcog & 55.41 & 58.21 & 52.50  & 53.12 & 55.31 \\
NVIDIA-AIQ & 56.90  & 58.49 & 52.89 & 53.43 & 55.95 \\
Grep Deep Research & 56.82 & 58.92 & 53.38 & 53.44 & 56.23 \\
Octen DeepResearch & 56.89 & 59.00    & 53.39 & 53.83 & 56.31 \\
1688AILab-DeepResearch & 57.32 & 59.27 & 53.51 & 53.36 & 56.53 \\
Cellcog-Max & 57.40  & 60.01 & 53.25 & 53.21 & 56.67 \\
Xiaoyi DeepResearch 6.0 & \underline{58.58} & 59.38 & 53.58 & 53.99 & 57.00 \\
Zhipu Deep Research & 58.15 & \underline{60.14} & 53.47 & 53.88 & 57.06 \\
iFlow-Researcher & 58.24 & 59.74 & 53.24 & \textbf{55.05} & 57.08 \\
ZTE Nebula DeepResearch & 58.37 & 59.76 & \textbf{54.06} & \underline{54.66} & \underline{57.27} \\
\midrule
\textbf{DuMate-DeepResearch} & \textbf{59.48} & \textbf{61.48} & \underline{53.87} & 54.34 & \textbf{58.03} \\
\bottomrule
\end{tabular}
}
\caption{Performance of different deep research models/systems on the DeepResearch Bench. The scores are presented in percentage, and the best and second-best performances are highlighted in bold and underline, respectively. The models/systems marked with an * represent results reproduced by the DeepResearch Bench paper. We report the performance of DuMate-DeepResearch based on average scores across multiple runs.}
\label{tab:deepresearch_bench_results}
\end{table}

\paragraph{DeepResearch Bench}We report the results of our DuMate-DeepResearch system and baselines on the DeepResearch Bench in Table~\ref{tab:deepresearch_bench_results}.
Table~\ref{tab:deepresearch_bench_results} demonstrates that DuMate-DeepResearch achieves the best overall score of 58.03\%, outperforming the second-best ZTE Nebula DeepResearch (57.27\%).
As for the individual evaluation dimensions, DuMate-DeepResearch ranks first in both Comprehensiveness (59.48\%) and Insight (61.48\%), improving over the second-best system by 0.90\% and 1.34\%, respectively. It ranks second on Instruction Following (53.87\%) and remains highly competitive on Readability (54.34\%), staying within 0.2--0.7\% of the top systems on these two dimensions. These results indicate that DuMate-DeepResearch can effectively acquire and synthesize information during the deep research process, and generate high-quality reports that are comprehensive, insightful, and well-structured.

\paragraph{DeepResearch Bench II}
We further evaluate on DeepResearch Bench II, which diagnoses deep research agents via fine-grained binary rubrics derived from expert reports. The benchmark assesses three dimensions: \textit{Information Recall} (whether the system retrieves all key facts), \textit{Analysis} (whether the system performs correct reasoning and synthesis), and \textit{Presentation} (whether the report is well-structured and readable). Results are reported in Table~\ref{tab:deepresearch_bench_ii_results}.

\begin{table}[ht]
\centering
\resizebox{\textwidth}{!}{
\begin{tabular}{ccccc}
\toprule
Model/System & \textsc{Information Recall} & \textsc{Analysis} & \textsc{Presentation} & \textsc{Overall} \\
\midrule
Tongyi Deep Research & 22.95 & 35.89 & 86.13 & 29.89 \\
Perplexity Research & 33.05 & 44.47 & 79.34 & 38.58 \\
Grok Deep Search & 33.52 & 42.50 & 91.42 & 39.23 \\
Qwen3-Max Deep Research & 34.18 & 48.04 & 74.59 & 39.25 \\
Doubao Deep Research & 34.83 & 49.43 & 83.51 & 40.99 \\
Gemini-2.5-Pro Deep Research & 34.91 & 51.91 & 90.24 & 41.98 \\
Gemini-3-Pro Deep Research & 39.09 & 48.94 & 91.85 & 44.60 \\
OpenAI-GPT-o3 Deep Research & 39.98 & 49.85 & 89.16 & 45.40 \\
NVIDIA-AIQ & 49.23 & 61.55 & \textbf{93.15} & 54.50 \\
CMCC-DeepInsight & 49.60 & 62.95 & \underline{92.94} & 55.39 \\
Xiaoyi DeepResearch 6.0 & 53.05 & \underline{69.90} & 91.12 & 58.72 \\
iFlow-Researcher & \underline{54.99} & 69.54 & 92.56 & \underline{59.91} \\
\midrule
\textbf{DuMate-DeepResearch} & \textbf{57.58} & \textbf{71.70} & 89.89 & \textbf{61.95} \\
\bottomrule
\end{tabular}
}
\caption{Performance on DeepResearch Bench II. Scores are percentages. The best and second-best performances are highlighted in bold and underline, respectively.}
\label{tab:deepresearch_bench_ii_results}
\end{table}

Table~\ref{tab:deepresearch_bench_ii_results} shows that, under our evaluation on DeepResearch Bench II, DuMate-DeepResearch achieves the best overall score of 61.95\%, outperforming the strongest baseline iFlow-Researcher by 2.04\%. It also ranks first in Information Recall (57.58\%) and Analysis (71.70\%), improving over the second-best systems by 2.59\% and 1.80\%, respectively. The rubric-based evaluation indicates that our system excels particularly in acquiring key evidence and performing evidence-grounded synthesis---the two capabilities most directly impacted by our graph-based dynamic planning and multi-turn retrieval mechanisms---while maintaining competitive Presentation quality (89.89\%).

\subsection{Detailed Analysis}

\paragraph{Ablation Study} To understand the contribution of key design choices in DuMate-DeepResearch, we conduct ablation studies on DeepResearch Bench, examining the impact of rubric-guided generation and the choice of report-stage model. Average results from 3 runs are reported in Table~\ref{tab:ablation_results}.

\begin{table}[ht]
\centering
\resizebox{\textwidth}{!}{
\begin{tabular}{lccccc}
\toprule
Variant & \textsc{Comprehensiveness} & \textsc{Insight} & \textsc{Instruction Following} & \textsc{Readability} & \textsc{Overall} \\
\midrule
\textbf{DuMate-DeepResearch (Full)} & 59.48 & 61.48 & 53.87 & 54.34 & 58.03 \\
\midrule
\multicolumn{6}{l}{\textit{Rubric Ablation}} \\
\quad w/o Rubric (Report Stage) & 59.01 & 60.73 & 53.62 & 53.82 & 57.61 \\
\quad w/o Rubric (Full Pipeline) & 58.95 & 60.78 & 53.71 & 53.91 & 57.53 \\
\midrule
\multicolumn{6}{l}{\textit{Report-Stage Model Replacement}} \\
\quad DeepSeek V4 Pro & 58.73 & 60.66 & 53.53 & 52.64 & 57.21 \\
\quad GLM 5.1 & 57.92 & 60.02 & 52.93 & 53.93 & 56.69 \\
\quad MiniMax-M3 & 55.91 & 58.75 & 51.75 & 51.64 & 55.21 \\
\quad Qwen-3.7 Max & 56.20 & 58.48 & 52.41 & 52.80 & 55.55 \\
\bottomrule
\end{tabular}
}
\caption{Ablation study results on DeepResearch Bench. ``w/o Rubric (Report Stage)'' removes rubric guidance only during report generation; ``w/o Rubric (Full Pipeline)'' removes rubric from all stages including planning and research. The report-stage model replacement variants substitute the default report generation model with the specified alternative while keeping all other components unchanged.}
\label{tab:ablation_results}
\end{table}

\paragraph{Effect of Rubric Guidance}
Removing the rubric from the report stage alone causes a modest but consistent drop across all dimensions (Overall: 58.03$\to$57.61, $-$0.42), with the largest degradation on Insight ($-$0.75) and Comprehensiveness ($-$0.47). Notably, further removing the rubric from planning and research stages yields only marginal additional decline (Overall: 57.53, a further $-$0.08 over report-only removal). This asymmetry indicates that the rubric's primary value materializes during report synthesis---where it serves as a live scaffold for evidence-grounded claim generation---rather than during earlier information-gathering stages. The finding aligns with our design intent (Section~\ref{sec:rubric}): persistent rubrics condition the Writer policy to ground claims in retrieved evidence at generation time, and this conditioning effect dominates the rubric's contribution to overall quality.

\paragraph{Effect of Report-Stage Model}
Replacing the default report-generation model produces substantially larger quality differences than rubric removal, confirming that the synthesis model is the single most impactful component in the pipeline. DeepSeek V4 Pro comes closest to the full system ($-$0.82 overall) but exhibits a notable Readability deficit ($-$1.70), suggesting weaker long-form formatting and structural coherence despite competitive analytical ability. GLM 5.1 maintains strong Readability (53.93, only $-$0.41) yet shows marked drops in Comprehensiveness ($-$1.56) and Insight ($-$1.46), indicating difficulty in fully leveraging the retrieved evidence base. MiniMax-M3 and Qwen-3.7 Max incur the largest overall degradations ($-$2.82 and $-$2.48, respectively), with broad declines across all dimensions; both models appear to struggle with the long-context, multi-source synthesis demands of deep research reports. Across all substitutions, the strongest models preserve Insight more robustly than Comprehensiveness, suggesting that information coverage---assembling all relevant evidence into a coherent narrative---is particularly sensitive to model capability and benefits most from scale.

\subsection{Qualitative Case Study}
\label{sec:case_study}

\subsubsection{Coarse-to-Fine Expansion and Dynamic Boundary Definition}

\textbf{Case~A:} ``How do low-code/no-code platforms impact traditional software development?'' This ambiguous query embeds four interleaved sub-problems: impact magnitude, efficiency vs.\ maintenance cost, developer vs.\ business perspectives, and future trends. Rather than immediately committing to fine-grained investigation, the system executes a two-phase expansion strategy.

\paragraph{Coarse Phase.} The \textit{initial\_planner} (the Planner's first-stage coarse expansion) issues two parallel exploratory search tasks to map the macro landscape, followed by an outline generation task (executed by the Writer) that depends on both:

\begin{small}
\begin{verbatim}
"task_graph": [
  {"subtask_id":"T-1", "subtask_type":"search",
   "subtask_title":"LCNC market status and impact on SDLC",
   "subtask_dependencies":[], "subtask_depth":1},
  {"subtask_id":"T-2", "subtask_type":"search",
   "subtask_title":"Efficiency gains vs. maintenance costs:
    empirical evidence and controversies",
   "subtask_dependencies":[], "subtask_depth":1},
  {"subtask_id":"T-3", "subtask_type":"outline",
   "subtask_title":"Generate structured research outline",
   "subtask_dependencies":["T-1","T-2"], "subtask_depth":2}
]
\end{verbatim}
\end{small}

\paragraph{Fine Phase.} Upon completion of T-1 and T-2, the Writer synthesizes an 8-chapter structured outline covering background, restructuring mechanisms, efficiency verification, hidden costs, stakeholder perspectives, platform comparison, boundaries, and future trends. This outline then triggers the \textit{planner} to expand the research into 14 targeted subtasks (T-4 through T-17) across three depth layers:

\begin{small}
\begin{verbatim}
Depth-1 (parallel): T-4..T-13 (10 search tasks)
  - Market background, traditional dev pain points,
    6-dimension restructuring, efficiency data,
    hidden costs, stakeholder views, platform comparison,
    industry cases, capability boundaries, future trends
Depth-2 (dependent): T-14 (llm), T-15 (llm), T-16 (search)
  - T-14: Cross-validate efficiency vs. cost data
  - T-15: Build scenario-platform matching matrix
  - T-16: Supplement opposing viewpoints
Depth-3: T-17 (report) [deps: T-4..T-16]
\end{verbatim}
\end{small}

\begin{figure}[t]
\centering
\begin{tikzpicture}[
    node distance=0.4cm and 0.6cm,
    every node/.style={font=\scriptsize},
    phase/.style={draw, rounded corners, fill=blue!8, minimum width=1.8cm, minimum height=0.5cm, align=center},
    task/.style={draw, rounded corners, fill=green!10, minimum width=1.2cm, minimum height=0.4cm, align=center},
    arrow/.style={->, >=stealth, thick}
]
\node[phase] (query) {User Query\\(4 sub-problems)};
\node[phase, below=0.5cm of query] (router) {Router};
\node[phase, below=0.5cm of router] (planner1) {Planner\\(coarse planning)};
\node[task, right=1.2cm of planner1] (t12) {T-1\&T-2\\search};
\node[task, below=0.5cm of t12] (outline) {\ \ T-3\ \ \\outline};
\node[phase, below=1.8cm of planner1] (planner2) {Planner\\(fine expansion)};
\node[task, below left=0.5cm and 1.2cm of planner2] (d1) {T-4..T-13\\10$\times$search};
\node[task, below=0.5cm of planner2] (d2) {T-14..T-16\\2$\times$llm + 1$\times$search};
\node[task, below right=0.5cm and 1.2cm of planner2] (d3) {T-17\\report};

\draw[arrow] (query) -- (router);
\draw[arrow] (router) -- (planner1);
\draw[arrow] (planner1) -- (t12);
\draw[arrow] (t12) -- (outline);
\draw[arrow] (planner1) -- (planner2);
\draw[arrow] (planner1) -- (outline);
\draw[arrow] (planner2) -- (d1);
\draw[arrow] (planner2) -- (d2);
\draw[arrow] (planner2) -- (d3);
\draw[arrow] (d1) -- (d2);
\draw[arrow] (d2) -- (d3);

\node[right=0.1cm of outline, blue!60] {\textbf{Coarse}};
\node[left=0.1cm of d1, red!60] {\textbf{Fine}};
\end{tikzpicture}
\caption{Coarse-to-fine expansion in Case~A. The coarse phase (router $\to$ planner $\to$ 2 searches $\to$ outline generation) establishes research boundaries; the fine phase (14 subtasks across 3 depth layers) performs targeted investigation.}
\label{fig:coarse_to_fine}
\end{figure}
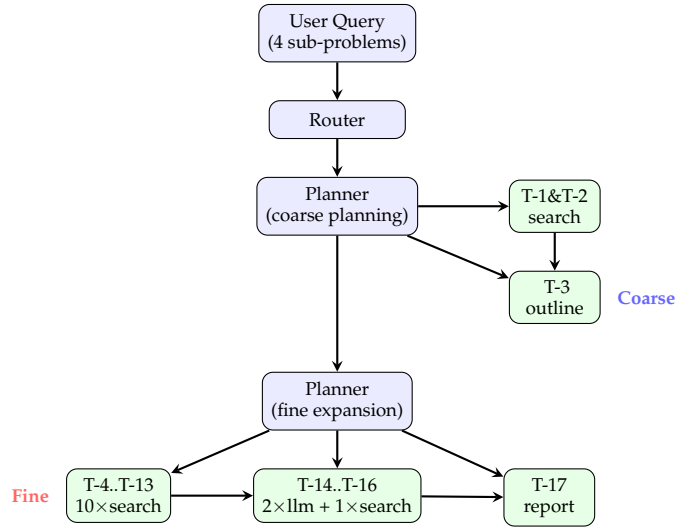

Figure~\ref{fig:coarse_to_fine} illustrates this two-phase expansion. The coarse phase establishes cognitive boundaries (``what is the research space?'') before the fine phase commits computational resources to depth-first investigation.

\subsubsection{Graph-Based Dynamic Planning and Reflection}

\begin{figure}[t]
\centering
\begin{tikzpicture}[
    node distance=0.3cm and 0.15cm,
    every node/.style={font=\tiny},
    snode/.style={draw, circle, fill=green!15, minimum size=0.5cm, inner sep=1pt},
    lnode/.style={draw, circle, fill=orange!20, minimum size=0.5cm, inner sep=1pt},
    rnode/.style={draw, circle, fill=red!15, minimum size=0.5cm, inner sep=1pt},
    arrow/.style={->, >=stealth, thin}
]
\foreach \i in {1,...,11} {
    \node[snode] (s\i) at ({(\i-1)*0.78}, 0) {T-\i};
}
\node[lnode] (l12) at (0, -1.3) {T-12};
\node[lnode] (l13) at (2.34, -1.3) {T-13};
\node[lnode] (l14) at (4.29, -1.3) {T-14};
\node[lnode] (l15) at (5.46, -1.3) {T-15};
\node[lnode] (l16) at (5.46, -2.6) {T-16};
\node[snode] (s17) at (3.12, -2.6) {T-17};
\node[rnode] (r18) at (4.29, -3.8) {T-18};

\draw[arrow] (s1) -- (l12);
\draw[arrow] (s2) -- (l13);
\draw[arrow] (s3) -- (l13);
\draw[arrow] (s4) -- (l13);
\draw[arrow] (s5) -- (l13);
\draw[arrow] (s6) -- (l14);
\draw[arrow] (s7) -- (l14);
\draw[arrow] (s8) -- (l15);
\draw[arrow] (l12) -- (l16);
\draw[arrow] (l13) -- (l16);
\draw[arrow] (l14) -- (l16);
\draw[arrow] (l15) -- (l16);
\draw[arrow] (l13) -- (s17);
\draw[arrow] (l14) -- (s17);
\draw[arrow] (l15) -- (s17);
\draw[arrow] (s9) to[out=-90, in=0] (s17);
\draw[arrow] (s10) to[out=-60, in=45] (l16);
\draw[arrow] (s11) to[out=-45, in=30] (l16);
\draw[arrow] (l16) -- (r18);
\draw[arrow] (s17) -- (r18);

\node[snode, right=0.4cm of s11] (leg1) {};
\node[right=0.1cm of leg1] {search};
\node[lnode, below=0.15cm of leg1] (leg2) {};
\node[right=0.1cm of leg2] {llm};
\node[rnode, below=0.15cm of leg2] (leg3) {};
\node[right=0.1cm of leg3] {report};

\node[left=0.1cm of s1, gray] {d=1};
\node[left=0.1cm of l12, gray] {d=2};
\node[left=0.1cm of s17, gray] {d=3};
\node[left=0.1cm of r18, gray] {d=4};
\end{tikzpicture}
\caption{Task execution DAG for Case~B. Depth-1: 11 parallel search tasks; depth-2: 4 \texttt{llm} tasks for integration and cross-validation; depth-3: cross-dimension synthesis (\texttt{llm}, T-16) and gap-filling search (T-17); depth-4: final report. T-17 depends on the integration tasks (T-13--T-15) and recovers the sub-segment data deferred from T-9; T-10--T-11 skip depth-2 and feed directly into T-16.}
\label{fig:task_dag}
\end{figure}
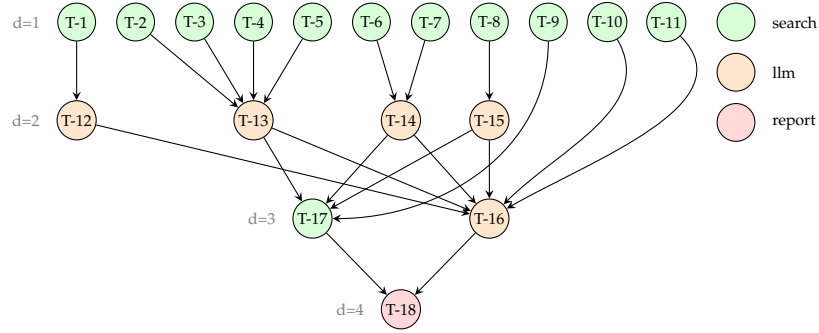

\textbf{Case~B:} ``Constructing a three-dimensional evaluation framework for NEV powertrain commercialization thresholds.'' The \textit{planner} constructs a four-layer DAG with 18 nodes, where edges encode strict execution dependencies (Figure~\ref{fig:task_dag}).

\paragraph{Reflective Evaluation.} At each scheduling cycle, the planner performs explicit quality assessment before deciding next actions:

\begin{small}
\begin{verbatim}
"last_task_revision":
  "T-1 (methodology): HIGH. Provides S-curve (10%/16%/50%),
   AHP+entropy weighting, TCO five-dimension framework.
   T-2 (800V+SiC): HIGH. Covers 40+ production models,
   substrate price curves, 5-10% efficiency gain data.
   T-3 (solid-state battery): HIGH. Covers three routes,
   350-500 Wh/kg density, 2025/2027/2030 milestones.
   ...
   T-9 (enterprise cases): ADEQUATE. Missing Hongqi/Lantu/
   Toyota-Mirai sub-segment data -- defer to T-17.
   T-11 (scenario forecasts): HIGH. Three-scenario matrix,
   BNEF/McKinsey/Ouyang cross-validated projections.
   Assessment: 11 search tasks complete, three-dimensional
   data coverage balanced. T-9 gap handled by T-17.
   Proceeding to T-12..T-15 (parallel llm integration)."
\end{verbatim}
\end{small}

This reflection-before-action loop enables the system to: (1)~confirm sufficient evidence before advancing to dependent tasks; (2)~dynamically inject additional searches when gaps are detected; and (3)~prune unnecessary branches when early results already satisfy requirements. The \texttt{llm}-type tasks (T-12--T-16) serve dedicated integration and cross-validation roles---synthesizing per-dimension indicators, computing composite scores, and verifying consistency across multiple search results rather than performing new searches.

\subsubsection{Multi-Turn Retrieval within Search Agents}

\textbf{Case~C:} ``Manufacturing technology options for hollow motor shafts in NEV electric drive units.'' Beyond planner-level re-planning, each search task executes a multi-turn retrieval loop internally. The \textit{Search Agent} operates as a plan-execute cycle with up to 10 iterations, progressively refining queries based on intermediate results.

A single search task (T-1) in this case executes 6 internal rounds with 40+ queries:

\begin{small}
\begin{verbatim}
Round 1 (broad): 3 search tools, 9 queries
  "hollow motor shaft NEV electric drive unit application"
  "hollow rotor shaft electric vehicle e-axle requirements"
  "hollow shaft rotor cooling 800V high speed motor NEV"

Round 2 (manufacturing-focused): 3 search tools, 9 queries
  "rotary swaging hollow rotor shaft EV production"
  "EV motor shaft material steel grade 42CrMo4 20MnCr5"
  "hairpin motor hollow shaft oil spray cooling rotor"

Round 3 (OEM-specific): 3 search tools, 9 queries
  "BYD 8-in-1 e-axle hollow rotor shaft 800V spec"
  "Tesla Model S Plaid drive unit hollow rotor shaft"
  "Hirschvogel multi-piece hollow rotor shaft laser welded"

Rounds 4-5: Progressively narrower (ISO standards, balance
  grades, specific tolerance specs)
Round 6: Final answer synthesis
\end{verbatim}
\end{small}

The multi-turn mechanism enables three retrieval strategies: (1)~\textbf{multi-formulation query expansion} (varying terminology, synonyms, and technical jargon to maximize recall); (2)~\textbf{progressive specificity} (broad domain $\to$ manufacturing process $\to$ OEM/supplier names $\to$ ISO standards); and (3)~\textbf{tool diversification} (search engine queries + direct URL crawling for authoritative sources).

\subsubsection{Rubric-Based Test-Time Optimization}

\begin{figure}[t]
\centering
\begin{tikzpicture}[
    node distance=0.5cm and 0.8cm,
    every node/.style={font=\scriptsize},
    box/.style={draw, rounded corners, minimum width=2cm, minimum height=0.6cm, align=center},
    rubric/.style={draw, dashed, rounded corners, fill=yellow!15, minimum width=2.5cm, minimum height=0.5cm, align=center},
    arrow/.style={->, >=stealth, thick},
    rubarrow/.style={->, >=stealth, thick, dashed, red!60}
]
\node[box, fill=orange!10] (planner1) {Planner\\(dispatch)};
\node[box, fill=blue!10, below=0.5cm of planner1] (woutline) {Writer\\(outline)};
\node[rubric, below=0.5cm of woutline] (prubric) {Persistent Rubrics\\(chapter descriptions)};
\node[box, fill=green!10, below left=0.8cm and 0.5cm of prubric] (search) {Search Agents\\(T-4..T-13)};
\node[box, fill=blue!10, below right=0.8cm and 0.5cm of prubric] (writer) {Writer\\(synthesis, T-17)};
\node[rubric, below=0.4cm of search] (erubric) {Ephemeral Rubrics\\(per-query criteria)};
\node[box, fill=orange!10, below=0.4cm of erubric] (planner2) {Planner\\(next-cycle)};

\draw[arrow] (planner1) -- (woutline);
\draw[arrow] (woutline) -- (prubric);
\draw[rubarrow] (prubric) -- (search) node[midway, left, red!60] {inject};
\draw[rubarrow] (prubric) -- (writer) node[midway, right, red!60] {inject};
\draw[arrow] (search) -- (erubric);
\draw[rubarrow] (erubric) -- (planner2) node[midway, right, red!60] {return};
\end{tikzpicture}
\caption{Rubric propagation in multi-agent collaboration. Persistent rubrics (from the Writer) are injected into both the Search Agents and the Writer. Ephemeral rubrics generated during search are returned to the Planner for next-cycle calibration.}
\label{fig:rubric_flow}
\end{figure}
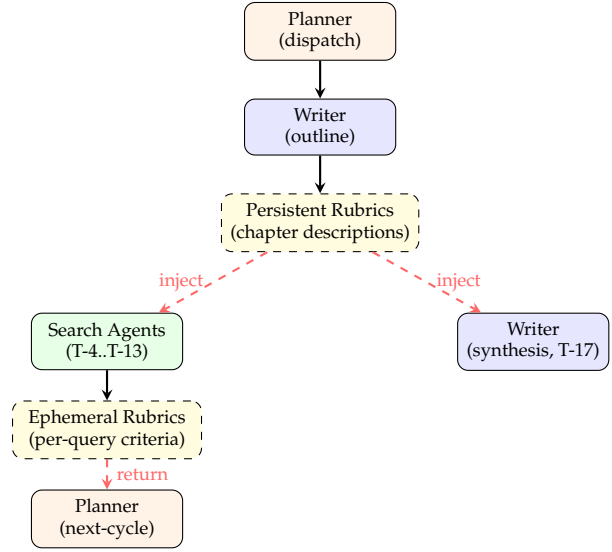

\textbf{Case~A:} ``How do low-code/no-code platforms impact traditional software development?'' The outline's chapter descriptions function as \textit{persistent rubrics} that scaffold all downstream agents. Figure~\ref{fig:rubric_flow} illustrates the rubric propagation pathway.

\paragraph{Rubric as Reasoning Scaffold.} Chapter~3's rubric specifies:

\begin{quote}
\small\itshape
``The study shall cross-validate LCNC efficiency claims using multi-source data: comparing delivery cycles, headcount, and ROI between vendor claims, third-party research (Forrester TEI, Gartner Peer Insights), and hands-on testing of 5--7 mainstream platforms; reveal the differentiated realization degree of efficiency gains across scenarios.''
\end{quote}

This rubric propagates to Search Agents (guiding query formulation toward multi-source evidence) and to the Writer (enforcing evidence grounding). The effect is directly observable in the final output, where the system produces \textbf{conditional, source-calibrated conclusions}:

\begin{quote}
\small\itshape
``The `300\%--500\% efficiency improvement' should be treated as the upper bound of vendor claims, not the median actually achievable by enterprises---this gap will be critically examined in Chapter~3. [...] IDC's 40.3B RMB (2024) with 26.4\% CAGR provides the most rigorous baseline; Gartner's 131B RMB figure includes broader aPaaS integration.''
\end{quote}

\subsubsection{Report Quality and Synthesis Capability}

Table~\ref{tab:case_quality} summarizes the output quality metrics across all three cases.

\begin{table}[h]
\centering
\small
\begin{tabular}{lccc}
\toprule
Metric & Case A (LCNC) & Case B (NEV) & Case C (Shaft) \\
\midrule
Word count\textsuperscript{\dag} & 151K (zh) & 261K (zh) &  68K (en) \\
Chapters / sections & 8 / 53 & 10 / 62 & 8 / 71 \\
Citations & 114 & 196 & 21 \\
Plan iterations & 2 & 3 & 11 \\
Total subtasks & 17 & 18 & 27 \\
\midrule
\multicolumn{4}{l}{\textit{Structural elements in final report:}} \\
Analytical frameworks & Mermaid, & AHP-entropy & Multi-criteria \\
 & matrices & formulas & decision matrix \\
Conditional conclusions & per-scenario & per-route & per-process \\
\bottomrule
\multicolumn{4}{l}{\textsuperscript{\dag}\scriptsize Chinese counts are in characters; English count is in words.} \\
\end{tabular}
\caption{Output quality metrics for all three case studies.}
\label{tab:case_quality}
\end{table}

All reports exhibit key quality characteristics enabled by the proposed mechanisms: (1)~\textbf{Multi-source cross-validation}: the system explicitly distinguishes vendor claims from third-party measurements (e.g., ``Forrester TEI validates 45\% cost reduction---notably more conservative than vendor-claimed 60--80\%''); (2)~\textbf{Conditional conclusions}: every major finding is bounded by scenario applicability (e.g., ``efficiency gains of 500--600\% in simple form/approval scenarios, but only 60\% in high-complexity projects''); (3)~\textbf{Quantitative modeling}: Case~B autonomously constructs a three-dimensional, 13-indicator evaluation framework with explicit formulas ($Score_k = \sum_i W_i \times \sum_j W_{ij} \times x_{ij,k}^{norm}$) and combined AHP-entropy weighting; (4)~\textbf{Adaptive depth}: Case~C demonstrates that the system scales plan iterations to 11 and total subtasks to 27 in response to retrieval difficulty, while maintaining report quality; (5)~\textbf{Full citation trails}: most evidence-backed claims link to a retrievable URL, enabling broad auditability.

These qualitative observations align with the quantitative gains on DeepResearch Bench, particularly the leading performance in Comprehensiveness (59.48\%, +0.9\% over second-best) and Insight (61.48\%, +1.34\% over second-best), which directly reflect the system's ability to acquire diverse evidence and synthesize it into structured, evidence-grounded analysis.

\section{Background and Related Work}

\subsection{Retrieval-Augmented Generation and Agentic Search}

Before deep research systems, the dominant paradigm for connecting LLMs with external knowledge was retrieval-augmented generation (RAG), where a system retrieves a small set of relevant passages and conditions the generator on them to produce a concise answer. Early RAG-style systems showed that non-parametric retrieval can substantially improve knowledge-intensive generation~\citep{lewis_etal_2020_retrieval}, and later work further integrated retrieval into language model pre-training and few-shot learning~\citep{guu2020retrieval,borgeaud2022improving,izacard2022few}. In these systems, the search component is usually optimized for short-answer question answering: retrieve evidence, optionally rerank or filter it, and generate an answer grounded in the retrieved context. The retriever itself has evolved from lexical retrieval such as BM25~\citep{robertson2009bm25} to dense passage retrieval~\citep{karpukhin2020dense}, while broader RAG surveys summarize this line as a standard way to mitigate the static-knowledge limitation of LLMs~\citep{gao_etal_2023_retrieval}.

A central limitation of conventional RAG is that retrieval quality depends heavily on the input query. To address this, many systems introduce LLM-based query rewriting, decomposition, or planning before retrieval~\citep{li2025towards,chen2025multi,li2026retain}. Rewrite-Retrieve-Read trains a query rewriter with reinforcement learning so that the rewritten query improves downstream answer accuracy~\citep{ma2023query,chen2026reflectrag}. Subsequent work extends this idea by optimizing retrieval-oriented planning with richer reward signals or multi-agent training, such as DeepRetrieval and multi-agent RAG optimization~\citep{jiang2025deepretrieval,chen2025improving}. Beyond one-shot rewriting, iterative systems decompose complex questions into multiple dependent sub-queries. LLatrieval repeatedly generates supplementary queries when current evidence fails verification~\citep{li2023llatrieval}, while DRAGIN uses the model's generation state to dynamically reformulate retrieval queries~\citep{su2024dragin}. Tree- or graph-based methods further expand the search space: RAG-Star uses retrieval-augmented verification and refinement over deliberative reasoning paths~\citep{jiang2024rag}, DeepRAG decides step by step whether to rely on parametric knowledge or retrieval~\citep{guan2025deeprag}, and MAO-ARAG orchestrates multiple retrieval modules through a multi-agent adaptive RAG framework~\citep{chen2025mao}.

Another line of work focuses on \textit{when} LLMs should search. Fixed retrieval can be inefficient and may introduce irrelevant or misleading evidence, so adaptive retrieval methods let the model decide whether additional evidence is needed. IR-CoT interleaves retrieval with chain-of-thought reasoning for multi-step questions~\citep{trivedi2022interleaving}, while FLARE triggers retrieval based on uncertainty during generation~\citep{jiang2023active}. Self-RAG trains models to retrieve, generate, and critique their outputs through self-reflection tokens~\citep{asai2024self}. Other adaptive methods estimate retrieval necessity through model confidence, internal states, or consistency, including DRAGIN, Rowen, and SEAKR~\citep{su2024dragin,ding2024retrieve,yao2024seakr}. This direction connects naturally to tool-using agents: ReAct frames search as an action interleaved with reasoning~\citep{yao_etal_2023_react}, Search-o1 introduces agentic search for large reasoning models~\citep{li2025search}, and Search-R1/R1-Searcher optimize when and what to search through reinforcement learning~\citep{jin2025search,song2025r1}.

Overall, LLM-augmented search and agentic RAG form the short-answer foundation of deep research. They improve evidence acquisition through retrieval, query planning, adaptive search timing, and tool-augmented reasoning. However, their primary objective is still usually localized answer accuracy or multi-hop question answering efficiency. Deep research extends this foundation from short, evidence-grounded answers to long-form, report-level synthesis, requiring broader tool orchestration, persistent memory, global planning, source calibration, and structured report generation.

\subsection{Deep Research}

Moving beyond short-answer RAG and agentic search, recent deep research systems aim to generate long-form, evidence-grounded reports for complex and open-ended user queries. Compared with conventional RAG systems, they usually require broader information exploration, longer-horizon planning, iterative reflection, source-level verification, and structured report writing. Therefore, the core challenge shifts from retrieving sufficient evidence for a localized answer to coordinating an end-to-end research workflow that can acquire, organize, and synthesize information across multiple steps.

MiroThinker~\citep{miromind_2025_mirothinker} is designed to enhance the tool-augmented reasoning ability and information-seeking capabilities of research agents. Operating on the ReAct~\citep{yao_etal_2023_react} paradigm, it supports up to 600 tool calls within a 256K context window by retaining the most recent tool responses during exploration. 
WebThinker~\citep{li_etal_2025_webthink_erempowering} introduces autonomous deep web exploration and operates in problem solving mode and report generation mode.
DR-Tulu~\citep{shao_etal_2025_drtulu_reinforcementlearning} addresses the drawback of static evaluation metrics in optimizing open-ended and long-form deep research tasks by introducing evolving rubrics. Rubrics provide measurable reward signals for RL and adapt dynamically to the policy model's behaviors.
TTD-DR~\citep{han_etal_2025_deepresearcher_test_time_diffusion} conceptualizes report generation as an iterative diffusion process, which includes planning, drafting, revision, and supplementary search. To enhance the quality of individual agentic components, TTD-DR introduces a self-evolution strategy that merges multiple revised variants into a single high-quality output.
Step-DeepResearch~\citep{hu_etal_2025_step_deepresearch_technical_report} adopts an Atomic Capability-based Data Synthesis Strategy for fine-tuning. The strategy targets several bottlenecks in deep research systems, including planning, information seeking, reflection, and report writing. Before SFT and RL, it introduces Agentic Mid-training to adapt medium-sized models to long-context and tool-augmented reasoning.
FS-Researcher~\citep{zhu_etal_2026_fsresearcher_testtimescaling_longhorizon} builds the research task as the collaboration between two agents: context builder and report writer. The system maintains a file-system workspace, which serves as the durable external memory for both agents. The context builder performs tool calls and knowledge base construction, while the report writer interacts with the file system and writes from section to section.

More recent systems further emphasize verification, scalable training data, and efficient long-horizon search. MiroThinker-1.7 and H1~\citep{team2026mirothinker} improve heavy-duty research agents through verification-enhanced data construction, scalable reinforcement learning, and inference-time verification. Marco DeepResearch~\citep{zhu2026marco} similarly adopts a verification-centric design, using a dedicated verification agent and reinforcement learning for compact models. RedSearch~\citep{chu2026redsearcher} targets scalable and cost-efficient long-horizon search agents by combining decentralized multi-agent data synthesis, compact agentic supervised fine-tuning, and reinforcement learning. LiteResearcher~\citep{li2026literesearcher} also focuses on scalable agentic RL for deep research, highlighting the importance of efficient trajectory generation and policy optimization.

Another emerging direction is to democratize deep research agents through open data and reproducible pipelines. OpenSeeker~\citep{du2026openseeker} fully open-sources its training data for frontier search agents, covering prompt sets, cold-start trajectories, and reinforcement learning data. OpenResearcher~\citep{li2026openresearcher} proposes a fully open pipeline for long-horizon deep research trajectory synthesis, including synthetic task generation, high-quality trajectory construction, and agent tuning. OffSeeker~\citep{zhou2026offseeker} argues that online reinforcement learning is not the only path to strong deep research agents, showing the effectiveness of offline data construction and training. AgentFounder~\citep{su2025scaling} scales agents through continual pre-training over large-scale agentic data, while DR-Venus~\citep{team2026dr} explores edge-scale deep research agents trained from only 10K open data examples.

Overall, existing deep research systems highlight several complementary directions: scaling tool-augmented exploration, separating problem-solving and report-generation modes, using rubrics and verifiers as optimization signals, improving test-time writing through iterative refinement, synthesizing capability-specific training data, open-sourcing reproducible training pipelines, and introducing external workspaces as persistent memory. These studies demonstrate that deep research is not merely a longer version of RAG, but a broader agentic workflow that couples search, planning, verification, memory, and long-form synthesis.
\section{Conclusions}

In this technical report, we presented \textbf{DuMate-DeepResearch}, a multi-agent deep research framework built on the Qianfan Agent Foundry. By decoupling the Agent Core, which handles task understanding, planning, and scheduling, from an extensible Tool Ecosystem for retrieval, evidence acquisition, and report rendering, the framework exposes every planning decision and tool invocation as an inspectable artifact, directly addressing the transparency and auditability challenge of agentic deep research. On top of this infrastructure, we introduced three cognitive mechanisms tailored to the open challenges of the task: a graph-based dynamic planner that supports coarse-to-fine exploration, reflection, re-planning, backtracking, and parallel branching for far-sighted long-horizon research; a recursive two-level execution design that delegates each complex search sub-task to an inner Search Agent running its own planning loop, isolating noisy retrieval so that the global trajectory stays stable; and a rubric-based test-time optimization mechanism that dynamically generates task-specific quality criteria and uses them as live reasoning scaffolds for evidence-grounded synthesis and adaptive stopping.

Experiments on DeepResearch Bench and DeepResearch Bench II show consistent gains across complementary evaluation protocols, with DuMate-DeepResearch achieving the best overall scores on both benchmarks. These results demonstrate the effectiveness of combining auditable multi-agent infrastructure with adaptive planning and rubric-guided reasoning for high-quality deep research. In future work, we plan to extend the evaluation to additional live and multimodal deep research benchmarks, broaden the Tool Ecosystem with richer domain-specific capabilities, and further investigate rubric-based optimization as a training-time as well as test-time signal.

\section*{Contributions and Acknowledgments}

\noindent\textbf{Contributors:}
Lingyong Yan\textsuperscript{\Letter},
Can Xu\textsuperscript{*},
Yukun Zhao,
Wenxuan Li,
Qingyang Chen,
Jiulong Wu,
Wenli Song,
Xiangnan Li,
Weixian Shi,
Yiqun Chen\textsuperscript{*},
Xuchen Ma\textsuperscript{*},
Yuchen Li,
Jiashu Zhao,
Shuaiqiang Wang,
Jianmin Wu, 
and Dawei Yin.

\medskip
\noindent\textsuperscript{\Letter}\,Corresponding author:
\href{mailto:yanlingyong@baidu.com}{yanlingyong@baidu.com}.\\
\textsuperscript{*}\,Work done during an internship at Baidu AI Cloud.

\medskip
\noindent We would like to thank our colleagues at Baidu AI Cloud and across Baidu for their continuous support throughout this project. We are also grateful to the colleagues who participated in internal evaluations and provided valuable feedback that helped shape the design and improve the quality of the system. Finally, we thank the broader open-source and deep research community, whose benchmarks, baselines, and prior work have been instrumental in guiding our research and development efforts.
\bibliography{colm2026_conference}
\bibliographystyle{colm2026_conference}

\appendix
\section{Prompt Templates}
\label{app:prompts}

To make the cognitive mechanisms of Section~\ref{sec:dpto} concrete and reproducible, this appendix reproduces desensitized excerpts of the core prompts that drive them.
Due to product and safety constraints, we release only high-level control logic: the full output schemas, field-level definitions, tool list, and other sensitive engineering details are omitted (marked in-line by a bracketed ellipsis), while the reasoning logic and control structure are retained.

\subsection{Planner Prompt}
\label{app:planner-prompt}

The following desensitized excerpt corresponds to the graph-based dynamic planner of Section~\ref{sec:graphplan}. It maintains and updates the research DAG, enforces the structural constraints, governs when to re-plan, and emits the next batch of parallel actions.

\begin{tcolorbox}[breakable,
  colframe=black!55, colback=black!3,
  fonttitle=\bfseries\footnotesize, fontupper=\footnotesize,
  title={Desensitized Planner Prompt (Excerpt)}]
\begin{CJK}{UTF8}{gbsn}
\textbf{【角色与任务】} 你是深度研究规划专家。请根据【用户需求】、【研究报告大纲】、【上一步完整计划图】与【上一步执行结果】，维护并输出一个完整、可执行、可动态更新的研究计划图（子任务组成的有向无环图 DAG）。核心职责：(1) 设计围绕需求与大纲的完整计划图；(2) 结合历史规划与执行结果，评估当前计划是否需要更新；(3) 确保任务覆盖关键问题、深度充分、依赖合理；(4) 输出下一步可并行执行的行动项。

\medskip
\textbf{【推理指导标准（Rubric）】} 本研究遵循注入的持久与单步 Rubric，所有规划决策都应参照这些标准：新增任务须满足标准要求；已执行结果若不达标，应规划补充或验证任务；任务目标应对齐 Rubric 的核心维度。〔此处注入的具体持久 / 单步 Rubric 内容略〕

\medskip
\textbf{【硬约束（节选）】} (1) 子任务须直接服务于需求与大纲，并显式对应具体章节；(2) 计划图须为合法 DAG，无循环依赖，同深度子任务互不依赖；(3) 规划总深度受上限约束，避免无意义的过深规划；(4) 子任务类型受工具列表限制；有且仅有一个报告类子任务并置于最终阶段，其余子任务不得承担最终结论或章节撰写；(5) 相对时间须基于当前时间转化为明确时间范围；(6) 输出须为合法的紧凑结构化结果。〔字段级约束与类型名称略〕

\medskip
\textbf{【重新规划策略】} 默认优先保持已有计划稳定，不为"看起来更全面"而随意扩展任务。仅当出现以下情形之一时才更新研究计划图：① 已执行任务失败、结果无效或明显偏题、无法支撑目标章节；② 关键章节尚未被覆盖；③ 已收集信息不足以支撑某些章节的深度要求；④ 存在信息冲突、不一致或不确定，需新增验证任务；⑤ 已执行结果暴露出高价值的新信息缺口，补充后能显著提升报告质量；⑥ 用户需求新增限制、目标变化或研究重点发生转移。若以上情形均不存在，应尽量复用既有未执行任务，不做无必要修改。

\medskip
\textbf{【任务设计规则】} \emph{通用规则}：每个子任务目标明确、可独立执行，避免过度碎片化，相近任务尽量合并；除报告类外，子任务仅负责信息收集、补充、验证与轻量整合校验，不得生成最终结论或章节内容。\emph{覆盖检查清单}（按相关性择优纳入，非机械全覆盖）：历史背景、当前状况、未来趋势、利益相关者、定量/定性证据、横向比较、风险与局限，以及主题特有维度。\emph{任务类型策略}：检索类用于获取与交叉验证外部信息；轻量推理类仅用于对已收集数据的去重、归并、统计与一致性校验，不得承担报告写作或综合分析；报告类唯一且置于最终阶段。

\medskip
\textbf{【输出规范】} 依次完成四步：(1) \emph{任务评估}——逐项评估上一步执行结果是否成功、达标、偏题或信息不足（无历史数据则按首次规划处理）；(2) \emph{计划决策}——说明本轮是否更新计划及具体动作（增 / 改 / 删任务或调整依赖）；(3) \emph{计划生成}——输出含已执行与未执行任务的完整计划图；(4) \emph{行动项生成}——选取依赖已满足、可立即并行执行的子任务，优先深度最小且信息增益最高者。〔字段定义、依赖与深度规则及 JSON 模板略〕
\end{CJK}
\end{tcolorbox}

\subsection{Rubric-Generation Prompts}
\label{app:rubric-prompt}

The rubric generator of Section~\ref{sec:rubric} operates at two levels. The \emph{orchestration-level} prompt (below) is invoked after each planning--execution cycle to assess cross-sub-task integration quality and to decide whether further retrieval is warranted; the \emph{search-level} prompt is invoked by each inner Search Agent after every tool response to steer its next retrieval step.

\begin{tcolorbox}[breakable,
  colframe=black!55, colback=black!3,
  fonttitle=\bfseries\footnotesize, fontupper=\footnotesize,
  title={Desensitized Orchestration-Level Rubric Prompt (Excerpt)}]
\begin{CJK}{UTF8}{gbsn}
\textbf{【角色】} 你是外层研究 Agent 的跨子任务信息整合质量标准生成器。外层 Agent 已收到多个内层子任务摘要，正在决策：信息是否足够支撑报告、是否需追加子任务、结果如何组织进大纲。请生成两类标准：贯穿后续决策的持久标准（Persistent）与基于当前汇总状态的单步标准（Ephemeral）。

\medskip
\textbf{【持久标准，四到六条】} 按主题类型从以下维度择优选取：① \emph{跨子任务覆盖完整性}（大纲各章节是否均有子任务覆盖，识别映射缺口）；② \emph{信息一致性}（摘要间是否存在矛盾，是否需交叉验证或取舍）；③ \emph{分析深度均衡性}（各章节深度是否均衡，有无过浅或冗余）；④ \emph{核心论点可支撑性}（结论是否有充分证据链，预测性结论是否含多情景或不确定性说明）；⑤ \emph{硬性指令完成度}（格式 / 数量 / 顺序等显式要求是否被集体满足）。每条 guidance 须面向"合并决策与大纲调整"给出可执行检查动作。

\medskip
\textbf{【单步标准，二到四条】} 对照大纲逐章节检查覆盖，识别三类跨子任务缺口（优先级递减）：\emph{覆盖空白}（章节无任何对应信息）、\emph{深度不均}（章节仅浅层信息）、\emph{冲突需裁决}（多子任务在同一维度信息不一致）。每条给出缺口类型、受影响章节与建议行动。

\medskip
\textbf{【关键约束】} 每条标准的 guidance 必须是可执行的推理指令，而非数字分数；当已收集信息已足够支撑报告撰写时，应显式标注"无需继续检索"，以向规划层提供自适应停止信号。〔持久 / 单步标准的完整 JSON 输出结构与字段细节略〕
\end{CJK}
\end{tcolorbox}

\begin{tcolorbox}[breakable,
  colframe=black!55, colback=black!3,
  fonttitle=\bfseries\footnotesize, fontupper=\footnotesize,
  title={Desensitized Search-Level Rubric Prompt (Excerpt)}]
\begin{CJK}{UTF8}{gbsn}
\textbf{【角色】} 你是内层深度搜索 Agent 的推理指导标准生成器。该 Agent 执行单个研究子任务，每次工具返回后请生成两类标准：持久标准（子任务全程有效，判断"信息是否值得保留 / 深入"）与单步标准（仅针对本次返回，指导"下一步检索什么"）。

\medskip
\textbf{【持久标准，三到五条】} 基于子任务目标，覆盖：\emph{相关性}（是否直接服务子任务目标）、\emph{具体性}（是否含可引用的数据、时间、来源、案例）、\emph{来源可靠性}（是否权威、有无时效性风险）、\emph{覆盖缺口感知}（哪些维度仍空白或浅层）、\emph{冲突识别}（与已有信息是否矛盾）。guidance 面向当前轮信息评估。

\medskip
\textbf{【单步标准，二到三条】} 对比"目标要求的信息"与"本次返回内容"，识别三类缺口（优先级递减）：\emph{缺失}（目标要求但本次未涉及）、\emph{深度不足}（仅概念级、缺数据 / 机制 / 案例）、\emph{延伸线索}（值得追踪的实体、时间或名称）。每条按"当前状态 / 搜索目标 / 建议行动"格式给出，并选取对下一步影响最大者。〔JSON 输出结构与字段细节略〕
\end{CJK}
\end{tcolorbox}

\end{document}